\documentclass[conference]{IEEEtran}
\usepackage{amsmath, amssymb, graphicx, cite, tabularx, booktabs, algorithm, algpseudocode}

\usepackage{tikz}
\usepackage{pgf}
\usetikzlibrary{positioning} 
\usepackage{float}
\usepackage{fancyhdr}
\usepackage{geometry}

\title{Hierarchical Performance-Based Design Optimization Framework for Soft Grippers%
\thanks{This research was funded in whole, or in part, by the Luxembourg National Research Fund (FNR), COSAMOS Project, ref. IC22/IS/17432865/COSAMOS. For the purpose of open access, and in fulfilment of the obligations arising from the grant agreement, the author has applied a Creative Commons Attribution 4.0 International (CC BY 4.0) license to any  Author Accepted Manuscript version arising from this submission.}}

\author{\IEEEauthorblockN{Hamed Rahimi Nohooji}
\IEEEauthorblockA{\textit{Automation and Robotics Research Group} \\
\textit{Interdisciplinary Centre for Security, Reliability and Trust}\\
\textit{University of Luxembourg, Luxembourg}\\
ORCID:0000-0001-9429-7164}
\and
\IEEEauthorblockN{Holger Voos}
\IEEEauthorblockA{\textit{Automation and Robotics Research Group} \\
\textit{Interdisciplinary Centre for Security, Reliability and Trust}\\
\textit{University of Luxembourg, Luxembourg}\\
ORCID:0000-0002-9600-8386}}

\fancypagestyle{plain}{%
  \fancyhf{}
  
  \geometry{top=54pt, bottom=54pt, left=54pt, right=54pt}
}

\begin{document}

\IEEEoverridecommandlockouts
\maketitle

\begin{abstract}
This paper presents a hierarchical, performance-based framework for the design optimization of multi-fingered soft grippers. To address the need for systematically defined performance indices, the framework employs a multi-objective optimization approach, structuring the optimization process into three integrated layers: Task Space, Motion Space, and Design Space. In the Task Space, performance indices are defined as core objectives, while the Motion Space interprets these into specific movement primitives. Finally, the Design Space applies parametric, topological, and field optimization techniques to refine the geometry and material distribution of the system, achieving a balanced design across key performance metrics. The framework’s layered structure enhances SG design, balancing competing design goals without predefined weight assignments, enabling flexible trade-off exploration and scalability for complex tasks.
\end{abstract}


\section{Introduction}

Soft robotic grippers (SGs) leverage the flexibility and compliance of soft actuators, offering a safe and adaptable solution for robotic manipulation across diverse environments. Unlike traditional rigid grippers, which are limited by their mechanical constraints, SGs leverage soft materials and compliant structures to handle objects of various shapes, sizes, and materials with precision and care, enabling complex tasks. This adaptability has positioned SGs as essential tools in a range of applications, from delicate handling in food processing to dynamic, unstructured object manipulation in fields like healthcare, wearables, and haptics, and robotics for human-machine interaction \cite{nohooji2024soft, shintake2018soft}.

Despite advancements in SGs, most current research centers on individual gripper configurations, primarily verified through experimental trials \cite{abozaid2024soft, qu2024advanced, kremer2023trigger}. While optimization is sometimes applied, these efforts are typically limited to single mechanical objectives, like enhancing motion or force distribution, rather than addressing SG performance in a holistic manner \cite{chen2020design, chen2022development, zhang2018design}. This narrow focus highlights the need for a systematic, performance-driven framework that balances key functional goals across a wide range of tasks, ensuring SGs are designed for versatile and reliable manipulation.

\begin{figure}[t]  
    \centering
    \begin{tikzpicture}[scale=0.85, x=0.75pt, y=0.75pt, yscale=-1, xscale=1]

\draw  [color={rgb, 255:red, 74; green, 144; blue, 226 }  ,draw opacity=1 ][fill={rgb, 255:red, 74; green, 144; blue, 226 }  ,fill opacity=0.21 ] (201.74,210.87) .. controls (201.74,173.11) and (232.35,142.5) .. (270.11,142.5) .. controls (307.87,142.5) and (338.48,173.11) .. (338.48,210.87) .. controls (338.48,248.63) and (307.87,279.24) .. (270.11,279.24) .. controls (232.35,279.24) and (201.74,248.63) .. (201.74,210.87) -- cycle ;
\draw  [color={rgb, 255:red, 65; green, 117; blue, 5 }  ,draw opacity=1 ][fill={rgb, 255:red, 126; green, 211; blue, 33 }  ,fill opacity=0.23 ] (213.8,223.93) .. controls (213.8,192.83) and (239.01,167.62) .. (270.11,167.62) .. controls (301.21,167.62) and (326.42,192.83) .. (326.42,223.93) .. controls (326.42,255.03) and (301.21,280.24) .. (270.11,280.24) .. controls (239.01,280.24) and (213.8,255.03) .. (213.8,223.93) -- cycle ;
\draw  [color={rgb, 255:red, 144; green, 19; blue, 254 }  ,draw opacity=1 ][fill={rgb, 255:red, 189; green, 16; blue, 224 }  ,fill opacity=0.28 ] (226.85,237.32) .. controls (226.85,213.67) and (246.02,194.5) .. (269.66,194.5) .. controls (293.31,194.5) and (312.48,213.67) .. (312.48,237.32) .. controls (312.48,260.96) and (293.31,280.13) .. (269.66,280.13) .. controls (246.02,280.13) and (226.85,260.96) .. (226.85,237.32) -- cycle ;
\draw  [color={rgb, 255:red, 144; green, 19; blue, 254 }  ,draw opacity=1 ][fill={rgb, 255:red, 189; green, 16; blue, 224 }  ,fill opacity=0.3 ] (72,238) .. controls (72,233.58) and (75.58,230) .. (80,230) -- (191,230) .. controls (195.42,230) and (199,233.58) .. (199,238) -- (199,262) .. controls (199,266.42) and (195.42,270) .. (191,270) -- (80,270) .. controls (75.58,270) and (72,266.42) .. (72,262) -- cycle ;
\draw  [color={rgb, 255:red, 65; green, 117; blue, 5 }  ,draw opacity=1 ][fill={rgb, 255:red, 184; green, 233; blue, 134 }  ,fill opacity=0.28 ] (341,194.09) .. controls (341,189.67) and (344.58,186.09) .. (349,186.09) -- (453,186.09) .. controls (457.42,186.09) and (461,189.67) .. (461,194.09) -- (461,218.09) .. controls (461,222.51) and (457.42,226.09) .. (453,226.09) -- (349,226.09) .. controls (344.58,226.09) and (341,222.51) .. (341,218.09) -- cycle ;
\draw  [color={rgb, 255:red, 144; green, 19; blue, 254 }  ,draw opacity=1 ][fill={rgb, 255:red, 189; green, 16; blue, 224 }  ,fill opacity=0.22 ] (341,238) .. controls (341,233.58) and (344.58,230) .. (349,230) -- (453,230) .. controls (457.42,230) and (461,233.58) .. (461,238) -- (461,262) .. controls (461,266.42) and (457.42,270) .. (453,270) -- (349,270) .. controls (344.58,270) and (341,266.42) .. (341,262) -- cycle ;
\draw  [color={rgb, 255:red, 74; green, 144; blue, 226 }  ,draw opacity=1 ][fill={rgb, 255:red, 74; green, 144; blue, 226 }  ,fill opacity=0.28 ] (72,151.09) .. controls (72,146.67) and (75.58,143.09) .. (80,143.09) -- (191,143.09) .. controls (195.42,143.09) and (199,146.67) .. (199,151.09) -- (199,175.09) .. controls (199,179.51) and (195.42,183.09) .. (191,183.09) -- (80,183.09) .. controls (75.58,183.09) and (72,179.51) .. (72,175.09) -- cycle ;
\draw  [color={rgb, 255:red, 65; green, 117; blue, 5 }  ,draw opacity=1 ][fill={rgb, 255:red, 184; green, 233; blue, 134 }  ,fill opacity=0.38 ] (72,195.09) .. controls (72,190.67) and (75.58,187.09) .. (80,187.09) -- (191,187.09) .. controls (195.42,187.09) and (199,190.67) .. (199,195.09) -- (199,219.09) .. controls (199,223.51) and (195.42,227.09) .. (191,227.09) -- (80,227.09) .. controls (75.58,227.09) and (72,223.51) .. (72,219.09) -- cycle ;
\draw  [color={rgb, 255:red, 74; green, 144; blue, 226 }  ,draw opacity=1 ][fill={rgb, 255:red, 74; green, 144; blue, 226 }  ,fill opacity=0.28 ] (341,150.09) .. controls (341,145.67) and (344.58,142.09) .. (349,142.09) -- (453,142.09) .. controls (457.42,142.09) and (461,145.67) .. (461,150.09) -- (461,174.09) .. controls (461,178.51) and (457.42,182.09) .. (453,182.09) -- (349,182.09) .. controls (344.58,182.09) and (341,178.51) .. (341,174.09) -- cycle ;
\draw  [color={rgb, 255:red, 245; green, 166; blue, 35 }  ,draw opacity=1 ][fill={rgb, 255:red, 245; green, 166; blue, 35 }  ,fill opacity=1 ] (152,173.45) -- (132.24,195.45) -- (112.48,173.45) -- (132.24,184.45) -- cycle ;
\draw  [color={rgb, 255:red, 245; green, 166; blue, 35 }  ,draw opacity=1 ][fill={rgb, 255:red, 245; green, 166; blue, 35 }  ,fill opacity=1 ] (152,217.45) -- (132.24,239.45) -- (112.48,217.45) -- (132.24,228.45) -- cycle ;
\draw  [color={rgb, 255:red, 245; green, 166; blue, 35 }  ,draw opacity=1 ][fill={rgb, 255:red, 245; green, 166; blue, 35 }  ,fill opacity=1 ] (81,282.82) .. controls (81,282.27) and (81.45,281.82) .. (82,281.82) -- (449.7,281.82) .. controls (450.25,281.82) and (450.7,282.27) .. (450.7,282.82) -- (450.7,285.82) .. controls (450.7,286.37) and (450.25,286.82) .. (449.7,286.82) -- (82,286.82) .. controls (81.45,286.82) and (81,286.37) .. (81,285.82) -- cycle ;

\draw (269.08,242.68) node [anchor=south] [inner sep=0.75pt]  [font=\footnotesize] [align=left] {\textbf{Design Space}};
\draw (239,152.09) node [anchor=north west][inner sep=0.75pt]  [font=\footnotesize] [align=left] {\textbf{Task Space}};
\draw (231,182.09) node [anchor=north west][inner sep=0.75pt]  [font=\footnotesize] [align=left] {\textbf{Motion Space}};
\draw (77,156.09) node [anchor=north west][inner sep=0.75pt]  [font=\footnotesize] [align=left] {High-Level Objective};
\draw (74,200.09) node [anchor=north west][inner sep=0.75pt]  [font=\footnotesize] [align=left] {Middle-Level Behavior};
\draw (72,244.09) node [anchor=north west][inner sep=0.75pt]  [font=\footnotesize] [align=left] {Low-Level Optimization};
\draw (341,156.09) node [anchor=north west][inner sep=0.75pt]  [font=\footnotesize] [align=left] {\textcolor[rgb]{0.82,0.01,0.11}{\textbf{Performance Indices }}};
\draw (348,191.09) node [anchor=north west][inner sep=0.75pt]  [font=\footnotesize] [align=left] {\begin{minipage}[lt]{71.96pt}\setlength\topsep{0pt}
\begin{center}
Bending \& Twisting
\end{center}
 Primitives
\end{minipage}};
\draw (347,236.09) node [anchor=north west][inner sep=0.75pt]  [font=\footnotesize] [align=left] {Optimization Model \\and Implementation};
\draw  [color={rgb, 255:red, 248; green, 231; blue, 28 }  ,draw opacity=1 ][fill={rgb, 255:red, 248; green, 231; blue, 28 }  ,fill opacity=1 ]  (121,119.82) -- (405,119.82) -- (405,137.82) -- (121,137.82) -- cycle  ;
\draw (124,123.82) node [anchor=north west][inner sep=0.75pt]  [font=\scriptsize] [align=left] {Optimizing Dexterity, Adaptability, Stability, and Reachability};

\end{tikzpicture}

\caption{A multi-layered framework for SGs optimization, integrating task-level objectives, motion behavior, and design-space refinement to improve dexterity, adaptability, stability, and reachability.}
    \label{fig:Arc}
\end{figure}
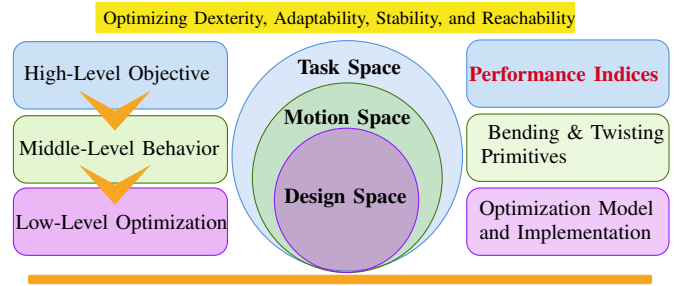

This paper proposes a performance-based design optimization framework to address this gap. The framework prioritizes an overarching goal of high-performance SGs, focusing on key performance indices (PIs)—adaptability, dexterity, reachability, and stability —that are essential for versatile manipulation. By structuring the optimization process around these indices, we employ a multi-objective optimization approach that systematically explores trade-offs among competing design goals without predefined weight assignments, ensuring flexibility in balancing diverse performance requirements.

To achieve this, we introduce a hierarchical design architecture comprising three interconnected levels as illustrated in Fig.\ref{fig:Arc}. At the highest level, the Task Space defines the primary performance objective, establishing the indices that guide the design. 
The Motion Space translates these indices into specific movement primitives, such as bending and twisting, which are critical for achieving practical SG performance. Finally, at the lowest level, the Design Space applies parametric, topological, and field optimization methods to refine the SG’s physical structure, ensuring that the design meets the targeted indices through an efficient and adaptable configuration of segments, actuator placements, and material distributions.

Through this structured approach, our framework systematically addresses the complex interactions between SG geometry, actuation, and task-specific requirements. By emphasizing performance as the foundation of design, this framework offers a pathway for advancing SG capabilities across a wide range of applications, enhancing their utility in scenarios requiring dexterity, stability, and adaptability.

The remainder of this paper is organized as follows: Section II defines the performance indices (PIs) and introduces the hierarchical architecture of the proposed framework. Section III presents the optimization model, integrating multi-objective goals aligned with the defined indices. Section IV details the implementation of parametric, topological, and field optimization methods, along with sensitivity analysis, to refine SG design. Section V discusses key findings, limitations, and future directions, concluding the paper.

\section{Optimization Framework for Soft Grippers}

\subsection{Problem Formulation}
In this study, we address the optimization of multi-fingered SGs by defining a structured design problem that considers the diverse, complex demands on SG performance. The SG design problem can be formulated as a multi-objective optimization task with competing requirements, necessitating a framework that can systematically incorporate and balance mechanical constraints, structural configurations, and actuation strategies. This formulation requires specifying design variables that directly impact SG geometry, material properties, and actuation mechanisms, as well as defining relevant constraints to ensure feasible operation under realistic load and interaction conditions. By framing the SG design as a structured optimization problem, this approach enables the integration of various performance objectives within a unified model, establishing a clear pathway for the comprehensive evaluation and improvement of SG functionality in complex applications.

\subsection{Optimization Goals}
\begin{figure}[t]
    \centering

\tikzset{every picture/.style={line width=0.75pt}} 

\begin{tikzpicture}[scale=0.81, x=0.75pt, y=0.75pt, yscale=-1, xscale=1]  

\draw  [color={rgb, 255:red, 255; green, 255; blue, 255 }  ,draw opacity=1 ][fill={rgb, 255:red, 255; green, 255; blue, 255 }  ,fill opacity=1 ] (143.33,188.79) -- (555.33,188.79) -- (555.33,516.79) -- (143.33,516.79) -- cycle ;
\draw  [color={rgb, 255:red, 248; green, 231; blue, 28 }  ,draw opacity=1 ][fill={rgb, 255:red, 248; green, 231; blue, 28 }  ,fill opacity=1 ] (147.77,290.27) .. controls (147.77,278.56) and (157.27,269.06) .. (168.99,269.06) -- (527.92,269.06) .. controls (539.64,269.06) and (549.14,278.56) .. (549.14,290.27) -- (549.14,383.79) .. controls (549.14,395.5) and (539.64,405) .. (527.92,405) -- (168.99,405) .. controls (157.27,405) and (147.77,395.5) .. (147.77,383.79) -- cycle ;
\draw  [color={rgb, 255:red, 128; green, 128; blue, 128 }  ,draw opacity=1 ][fill={rgb, 255:red, 245; green, 166; blue, 35 }  ,fill opacity=1 ][line width=2.25]  (266.33,337.06) .. controls (266.33,311.65) and (303.49,291.06) .. (349.33,291.06) .. controls (395.17,291.06) and (432.33,311.65) .. (432.33,337.06) .. controls (432.33,362.46) and (395.17,383.06) .. (349.33,383.06) .. controls (303.49,383.06) and (266.33,362.46) .. (266.33,337.06) -- cycle ;

\draw  [color={rgb, 255:red, 80; green, 227; blue, 194 }  ,draw opacity=0.6 ][fill={rgb, 255:red, 80; green, 227; blue, 194 }  ,fill opacity=0.6 ]  (352,160) -- (549,160) -- (549,264) -- (352,264) -- cycle  ;
\draw (355,164) node [anchor=north west][inner sep=0.75pt]   [align=left] {\textbf{Dexterity}\\{\small * }Fine manipulation control \ \ \\{\small * }Maximize precision in\\ \ \ complex tasks \ \ \ \ \ \ \ \ \ \ \ \ \ \ \ \ \ \ };
\draw  [color={rgb, 255:red, 74; green, 171; blue, 226 }  ,draw opacity=0.59 ][fill={rgb, 255:red, 74; green, 171; blue, 226 }  ,fill opacity=0.6 ]  (147,160) -- (347,160) -- (347,264) -- (147,264) -- cycle  ;
\draw (150,164) node [anchor=north west][inner sep=0.75pt]   [align=left] {\textbf{Adaptability}\\{\small * }Conform to object variety \ \ \ \\{\small  * }Enhance flexibility \\ \ \ while maintaining structure};
\draw  [color={rgb, 255:red, 189; green, 16; blue, 224 }  ,draw opacity=0.41 ][fill={rgb, 255:red, 189; green, 16; blue, 224 }  ,fill opacity=0.4 ]  (151,410) -- (348,410) -- (348,514) -- (151,514) -- cycle  ;
\draw (154,414) node [anchor=north west][inner sep=0.75pt]   [align=left] {\textbf{Reachability}\\{\small * }Extend operational range\\{\small * }Maximize reach \\ \ \ without losing control \ \ \ \ \ \ \ \ };
\draw  [color={rgb, 255:red, 184; green, 233; blue, 134 }  ,draw opacity=0.61 ][fill={rgb, 255:red, 184; green, 233; blue, 134 }  ,fill opacity=0.6 ]  (351,410) -- (548,410) -- (548,514) -- (351,514) -- cycle  ;
\draw (354,414) node [anchor=north west][inner sep=0.75pt]   [align=left] {\textbf{Stability}\\{\small * }Maintain secure grasp\\{\small  * }Ensure a robust grasp \ \ \ \ \ \ \\ \ \ \ while adapting \ \ \ \ \ \ \ \ \ \ \ \ \ \ \ \ \ };
\draw (276.17,310.33) node [anchor=north west][inner sep=0.75pt]  [font=\small] [align=left] {\begin{minipage}[lt]{89.2pt}\setlength\topsep{0pt}
\begin{center}
\textbf{Performance Indices}\\\textbf{in Design }\\\textbf{Optimization of SGs}
\end{center}

\end{minipage}};
\draw  [draw opacity=0]  (158.51,265) -- (327.51,265) -- (327.51,290) -- (158.51,290) -- cycle  ;
\draw (161.51,269) node [anchor=north west][inner sep=0.75pt]  [color={rgb, 255:red, 245; green, 105; blue, 35 }  ,opacity=1 ] [align=left] {{\small \textit{\textbf{Shape-conforming ability}}}};
\draw  [draw opacity=0]  (354.01,265) -- (541.01,265) -- (541.01,290) -- (354.01,290) -- cycle  ;
\draw (357.01,269) node [anchor=north west][inner sep=0.75pt]  [color={rgb, 255:red, 245; green, 105; blue, 35 }  ,opacity=1 ] [align=left] {{\small \textit{\textbf{Control over grasping tasks}}}};
\draw  [draw opacity=0]  (155.51,381) -- (345.51,381) -- (345.51,406) -- (155.51,406) -- cycle  ;
\draw (158.51,385) node [anchor=north west][inner sep=0.75pt]  [color={rgb, 255:red, 245; green, 105; blue, 35 }  ,opacity=1 ] [align=left] {{\small \textit{\textbf{Operational range extension}}}};
\draw  [draw opacity=0]  (367.14,381) -- (539.14,381) -- (539.14,406) -- (367.14,406) -- cycle  ;
\draw (453.14,385) node [anchor=north] [inner sep=0.75pt]  [font=\normalsize,color={rgb, 255:red, 245; green, 105; blue, 35 }  ,opacity=1 ] [align=left] {{\small \textit{\textbf{Resistant to disturbances}}}};

\end{tikzpicture}

    \caption{Performance indices for soft gripper optimization focusing on adaptability, dexterity, reachability, and stability. These indices guide the optimization process in relation to the gripper configuration.}
    \label{fig: PIs}
\end{figure}
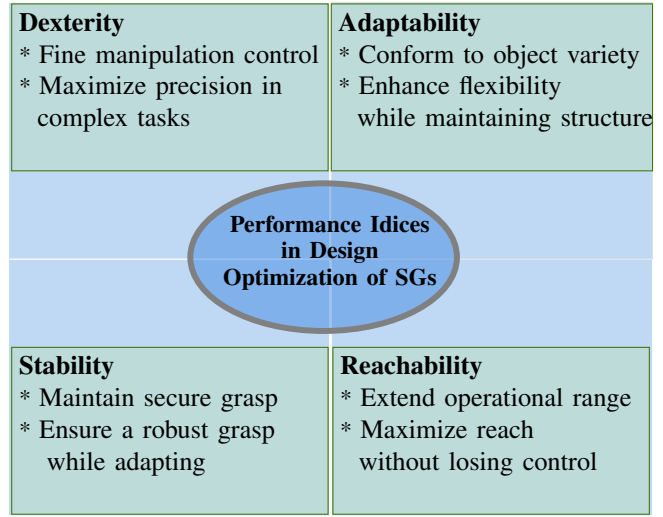

The primary aim of this framework is to systematically enhance SG performance by targeting key PIs. Each PI, illustrated in Fig. \ref{fig: PIs} and outlined below, functions as a measurable objective that guides the design towards robust and adaptable SGs, equipped for varied manipulation tasks and operating conditions. These PIs are then formulated into a multi-objective approach to ensure balanced optimization, driving overall effectiveness in diverse applications.

\begin{itemize}
\vspace{2pt}
    \item \textbf{Adaptability} relates to the gripper’s ability to conform to objects of varying shapes, sizes, and materials \cite{zhou2021bio}. For example, an SG handling irregularly shaped fruits adapts by deforming its structure to maintain a secure grasp. This index is crucial for ensuring that the gripper can handle a wide range of tasks without requiring reconfiguration. The design optimization process incorporates compliant chamber configurations and variable material properties, allowing the gripper to dynamically adjust its structure to accommodate different objects. In SG design, adaptability is improved by optimizing chamber layouts and material distribution to allow the gripper to deform effectively in response to objects, enhancing its ability to grasp and manipulate a diverse array of items. The optimization goal for adaptability is to ensure the gripper can maintain flexibility and increase compliance when adjusting to varying object properties without compromising its overall structural integrity.

      \item \textbf{Dexterity}, in the context of SGs, refers to the gripper’s ability to perform fine manipulation tasks with precision, handling objects \cite{abondance2020dexterous}. 
      For instance, a soft gripper assembling small electronic components requires precise joint articulation and finger coordination. The optimization process focuses on enhancing the geometric configuration of segments and chambers, ensuring smooth transitions in motion, and enabling precise control over various manipulation tasks. Dexterity is closely linked to the gripper’s segment length, chamber design, and joint articulation, with each of these variables contributing to the overall control and precision of the gripper’s movements. The design optimization goal for dexterity is to refine these geometric parameters to maximize the gripper’s ability to perform complex and delicate manipulation tasks while maintaining stability.
    
   \vspace{2pt}
    \item \textbf{Reachability} refers to the gripper’s capability to interact with objects across a broad workspace. For multi-fingered SGs, this index is directly influenced by the length and positioning of the segments, as well as the joint articulation.     
    Reachability involves extending the gripper’s operational range while maintaining control over the object. The optimization focuses on adjusting segment lengths and joint flexibility to ensure the gripper can reach objects at varying distances without sacrificing dexterity or stability.    
    The goal for reachability is to maximize the operational range of the gripper, enabling it to handle objects across different distances while maintaining control and performance in complex tasks.
\vspace{2pt}
     \item \textbf{Stability} is the gripper’s ability to maintain a secure grasp under dynamic conditions, such as external forces or unexpected movements \cite{xie2023learning}. In the context of SGs, stability is achieved by optimizing the distribution of forces across the gripper’s contact surfaces and ensuring robust material strength. The optimization process involves adjusting the gripper’s segment and chamber configuration to balance forces and minimize slippage, while also considering the material properties to prevent deformation under load. The goal of optimizing stability is to ensure the gripper can maintain a robust grasp on objects securely, even under unpredictable conditions, while preserving dexterity and adaptability.

\end{itemize}

\subsection{Hierarchical Architecture for Soft Gripper Design}

The hierarchical architecture for the design optimization of SGs is a structured approach that begins with a high-level performance objective and progressively refines it through intermediate indices and specific design parameters. Unlike traditional approaches that optimize SGs based solely on mechanical objectives (e.g., maximizing bending or grip force), this framework prioritizes performance as the overarching goal. To achieve this, the architecture systematically defines key PIs—dexterity, adaptability, stability, and reachability—interprets these indices into functional movement behaviors, and then refines the design parameters to meet these objectives (see Fig. \ref{fig:Arc}).

\vspace{10pt}

\textbf{Task Space (High-Level Performance Objective)}

At the highest level, the Task Space focuses on achieving a comprehensive performance objective designed for the versatile manipulation needs of SGs. Rather than maximizing isolated mechanical properties, this level establishes an overarching benchmark that SGs should aim to meet across diverse applications. Here, core PIs serve as structured guides, directing the optimization process to ensure balanced, reliable, and adaptable functionality.

Within the Task Space, PIs are defined based on the functional requirements of the SGs, such as handling objects of varied shapes, applying precise forces, and maintaining stable grasps under dynamic conditions. These indices establish the foundational benchmarks, setting the stage for a performance-driven approach that permeates all subsequent optimization stages in the framework.

Evaluation criteria for each PI are suggested according to specific task demands and expected functionality. These criteria provide a reference for guiding the optimization process, ensuring that each index is measurable and aligns with overarching performance objectives. In this way, the Task Space provides a structured pathway for systematically evaluating and enhancing SG performance in alignment with real-world requirements.

\vspace{10pt}

\textbf{Motion Space (Interpreting Indices into Movement Primitives)}
The Motion Space translates the high-level PIs into specific movement primitives and behaviors essential for SG functionality. This middle level focuses on bending, twisting, and controlled deformations—motions that directly contribute to achieving dexterity, adaptability, and stability in practical scenarios. By addressing these motion primitives, the Motion Space provides a bridge between the abstract PIs and the tangible movements the SG must perform.

Optimization within the Motion Space centers on configuring the SG’s segments and chambers to ensure smooth, coordinated actions aligned with the defined PIs. For example, bending and twisting are optimized to enhance adaptability, allowing the gripper to conform to a wide range of object geometries. Stability is reinforced by refining force distribution during these motions, ensuring secure handling even under varying loads. This level thus translates the desired performance attributes into operational behaviors that can be directly implemented in the design.

\vspace{10pt}

\textbf{Design Space (Low-Level Optimization of Parameters)}
At the lowest level, the Design Space applies parametric, topological, and field optimization techniques to refine the physical structure and material properties of the SG. This level focuses on adjusting specific design variables—such as segment lengths, chamber configurations, material elasticity, and actuator positioning—to meet the high-level PIs as expressed through the motion primitives defined in the Motion Space.

Parametric optimization in the Design Space involves tuning continuous geometric and material parameters to enhance flexibility, dexterity, and adaptability. Segment lengths, chamber configurations, and material properties are optimized to ensure smooth transitions in motion while maintaining structural integrity. Topological optimization focuses on the spatial distribution of material, enhancing stability and structural efficiency by concentrating material in high-stress areas while reducing it elsewhere. 
Field optimization introduces spatially varying design parameters, enabling localized control over stiffness, compliance, and actuation forces. By continuously adjusting these properties across the structure, this approach enhances adaptability and dexterity while maintaining structural integrity.
Together, these three optimization methods provide a comprehensive approach to refining the gripper’s design for balanced performance across key indices.

\begin{figure}
\centering
\includegraphics
[width=1\linewidth]{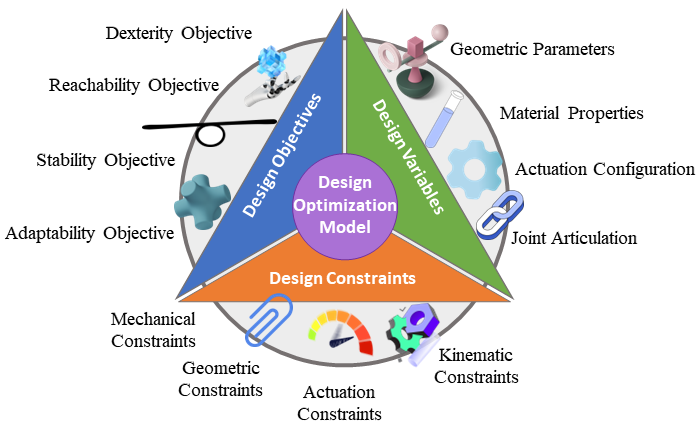}
\caption{Design Optimization Model for SGs, illustrating the key design objectives, variables, and constraints.}
\label{fig:opt model}
\end{figure}

\section{Optimization Model}

The optimization model forms the core of the design framework for SGs, aiming to systematically enhance their performance by maximizing key PIs. This model employs a multi-objective optimization approach, ensuring the identification of Pareto-optimal solutions, where no performance index can be improved without compromising another, thus balancing trade-offs effectively. By avoiding predefined weight assignments, this approach ensures a well-balanced design that remains adaptable across diverse manipulation tasks. The optimization model is composed of three main components—design objective, design variables, and constraints—each defined to align with the hierarchical structure of the framework, as shown in Fig. \ref{fig:opt model}.

\subsection{Design Objective}

The primary objective of the optimization model is to maximize the performance of the SG across the defined indices. Unlike traditional single-objective optimization  methods that require predefined weight assignments, this framework employs a multi-objective optimization  approach, systematically exploring trade-offs among competing performance indices. This ensures a balanced design without enforcing a fixed priority on any single objective, allowing flexibility in adapting to different manipulation requirements. The objective function is formulated as follows:
\fontsize{9.5pt}{12pt}
\begin{equation}
\min_{\mathbf{x}} \left[ f_{\text{dexterity}}(\mathbf{x}), f_{\text{adaptability}}(\mathbf{x}), f_{\text{stability}}(\mathbf{x}), f_{\text{reachability}}(\mathbf{x}) \right]
\end{equation}
\normalsize
where \( \mathbf{x} \) represents the vector of design variables. This approach enables the exploration of trade-offs and synergies between indices, providing a robust basis for optimizing SG performance in diverse scenarios. By avoiding explicit weight assignments, the model ensures that each objective is prioritized equally, thus promoting a balanced and adaptable design.

\subsection{Design Variables}

The design variables in this model are critical parameters that directly influence the PIs. These variables are chosen based on their impact on both the mechanical structure and functionality of the gripper, aligning with the lower-level design objectives in the hierarchical framework. Key design variables include:

\begin{itemize}
\vspace{5pt}
    \item \textbf{Geometric parameters}: These include segment lengths, actuation configurations, and actuator positioning. Adjusting these parameters affects the gripper’s dexterity and reachability, enabling precise control over movement and extension.
    \vspace{5pt}
    \item \textbf{Material properties}: Characteristics such as stiffness and elasticity directly influence the gripper’s adaptability and stability. By selecting appropriate materials or customizing their distribution, the gripper can achieve both structural integrity and flexibility, adapting effectively to varied object properties.
    \vspace{5pt}
    \item \textbf{Actuation configurations}: Actuation setups, such as pneumatic chamber layouts and tendon-driven mechanisms, are central to enabling the necessary motion primitives for bending, twisting, and reaching. Optimization of actuation configurations enhances adaptability and stability, ensuring that the gripper can respond dynamically to different manipulation tasks.

\vspace{5pt}
\item \textbf{Joint articulation}: Parameters related to joint flexibility, angles, and range of motion contribute significantly to dexterity and reachability. Optimizing joint articulation enables more nuanced control over the gripper's movement, allowing for precise manipulation across a broader workspace.
\end{itemize}

\subsection{Constraints}
The optimization process is subject to a range of constraints that ensure the design remains feasible and functional under real-world conditions. These constraints are defined to maintain the structural integrity of the gripper while allowing it to perform effectively within its intended operational space. The primary constraints are as follows:

\begin{itemize}
\vspace{5pt}
    \item \textbf{Mechanical constraints}: These constraints ensure that the gripper operates within safe stress and strain limits, preventing material failure or excessive deformation. Mechanical constraints include load-bearing capacity, maximum allowable stress, and limits on deformation, which are essential for maintaining stability under dynamic conditions.
    \vspace{5pt}
    \item \textbf{Geometric constraints}: These constraints define permissible ranges for segment lengths, chamber configurations, and actuator placements. They ensure that the gripper’s structure adheres to design specifications, allowing it to maintain reachability and dexterity without overextending or underutilizing its range.
    \vspace{5pt}
    \item \textbf{Actuation constraints}: These constraints regulate force and pressure limits within the actuation system, ensuring that the gripper performs consistently without overloading actuators. Actuation constraints are crucial for maintaining adaptability, as they allow the gripper to respond to varying object characteristics without losing control.
    \vspace{5pt}
    \item \textbf{Kinematic constraints}: Focusing on the gripper’s motion-related properties, kinematic constraints control joint angles, velocities, and movement ranges. These constraints ensure that each joint and segment operates within safe limits, preserving dexterity and reachability during complex manipulation tasks.
\end{itemize}

\section{Optimization Implementation}


\subsection{Optimization Formulation}

To formalize the optimization process, we define the following multi-objective formulation:

\begin{equation}
\begin{aligned}
    \min_{\mathbf{x}} & \quad \mathbf{F}(\mathbf{x}) = \left[ f_{\text{dext}}(\mathbf{x}), f_{\text{adapt}}(\mathbf{x}), f_{\text{stability}}(\mathbf{x}), f_{\text{reach}}(\mathbf{x}) \right] \\
    \text{s.t.} & \quad g_j(\mathbf{x}) \leq 0, \quad j = 1, \ldots, m \\
    & \quad h_k(\mathbf{x}) = 0, \quad k = 1, \ldots, p
\end{aligned}
\end{equation}
where \( \mathbf{x} \) represents the vector of design variables, incorporating geometric, material, and field-related parameters, ensuring a comprehensive optimization approach. The constraints \( g_j(\mathbf{x}) \) enforce mechanical feasibility, actuation limits, and spatial constraints. \( h_k(\mathbf{x}) \) ensures equality conditions such as boundary constraints and prescribed deformations.
This formulation allows for a balanced optimization process, integrating parametric, topological, and field optimization within a unified framework, enabling efficient design exploration without predefined weight assignments.

\subsection{Parametric Optimization}

Parametric optimization in this framework focuses on refining continuous design variables, including segment lengths, actuator configurations, joint articulation, and material properties, to enhance dexterity, adaptability, and reachability. For multi-fingered SGs, it enables precise tuning of segment dimensions and actuator placement to achieve smoother transitions and controlled deformations.

The objective function for parametric optimization considers variables that directly influence the movement and force distribution of each SG finger:
\begin{equation}
\max_{S, C} J(S, C) \quad \text{s.t.} \quad C_i(S, C) = 0, \, \forall i \in [1,k]
\end{equation}
where \( S \) represents segment lengths and articulations, \( C \) represents chamber design parameters (e.g., width, shape, and internal pressure capacity), and \( J(S, C) \) is a performance function combining indices.

For this optimization, gradient-based methods are effective for smooth problems, while heuristic approaches, such as genetic algorithms and particle swarm optimization, are more suitable when nonlinearity increases
\cite{ahmadianfar2020gradient}. By adjusting segment and chamber variables, the optimization identifies configurations that balance precise control over dexterous movement and the ability to adapt to diverse object geometries without rigid mechanical constraints. 
Moreover, sensitivity analysis can periodically apply within parametric optimization to determine the most impactful variables for each index, enabling more effective refinements as the process iterates.

\subsection{Topological Optimization}

Topological optimization focuses on optimizing the material distribution within each SG finger, ensuring structural efficiency across different actuation mechanisms. This approach is especially beneficial for SGs, where material properties, spatial distribution, and actuator placement influence force transmission, compliance, and stability under dynamic conditions.

The objective function for topological optimization can be expressed as:
\fontsize{9.5pt}{12pt}
\begin{equation}
\min_{\rho} F(\mathbf{U}(\rho), \rho) \quad \text{s.t.} \quad K(\rho)\mathbf{U} = \mathbf{P}, \quad \sum_{i=1}^{N} v_i \rho_i - V_f \leq 0
\end{equation}
\normalsize
where \( \rho \) represents the material density distribution, \( K(\rho) \) is the stiffness matrix dependent on the material density, \( \mathbf{U} \) represents displacement, and \( \mathbf{P} \) is the applied force vector. The term \( V_f \) sets the maximum volume fraction, imposing a constraint on material usage to ensure structural efficiency.

For material distribution, topological optimization uses density-based approaches like SIMP (Solid Isotropic Material with Penalization) \cite{zhang2018topology} and RAMP (Rational Approximation of Material Properties) \cite{chen2022development} to adjust density across the design domain, enhancing stability and strength. Level-set methods \cite{van2013level} offer an alternative by smoothly defining boundaries, creating clear material regions. Concentrating material in high-stress areas while reducing it in less critical ones preserves structural integrity, ensuring flexibility under load without compromising adaptability \cite{sigmund2013topology}. This balance allows for the integration of flexible and rigid materials, essential for precise bending and force application in complex tasks.

\subsection{Field Optimization}

Field optimization refines the spatial distribution of material properties, actuation forces, and compliance across the SG structure. 
Field optimization dynamically adjusts spatial material and actuation properties, allowing localized compliance control. This ensures adaptive force distribution and deformation behavior, improving grasp stability and response precision.
The objective function for field optimization is formulated as:
\begin{equation}
\min_{\mathbf{\phi}} F(\mathbf{U}(\mathbf{\phi}), \mathbf{\phi}) \quad \text{s.t.} \quad K(\mathbf{\phi})\mathbf{U} = \mathbf{P}
\end{equation}
where \( \mathbf{\phi} \) represents spatially varying design parameters such as stiffness gradients and actuator pressure fields, \( K(\mathbf{\phi}) \) is the stiffness matrix incorporating these variations, and \( \mathbf{U} \) represents the resulting displacement field.

Optimization techniques such as gradient-based methods, finite element-based sensitivity analysis, and physics-informed neural networks can be employed to iteratively refine the spatial distribution of parameters. By dynamically adjusting stiffness and actuation profiles, field optimization enhances adaptability and dexterity while maintaining structural integrity. This method ensures that soft grippers can locally modulate compliance, improving their ability to interact with objects of varying stiffness, shape, and fragility in real-time.

\subsection{Sensitivity Analysis}

Sensitivity analysis is to identify design variables that most significantly impact PIs. The sensitivity of an objective function \( f_i(\mathbf{x}) \) with respect to a design variable \( x_j \) is represented as:
\begin{equation}
S_{ij} = \frac{\partial f_i}{\partial x_j}
\end{equation}

This analysis targets the most influential variables, allowing refinement to focus on parameters critical to overall SG performance.
If initial optimization results do not meet design objectives, iterative adjustments are applied to these critical variables, ensuring refinements are both targeted and impactful for performance across tasks.

\subsection{Final Evaluation}
The optimized design undergoes a final evaluation to verify alignment with established performance criteria. This step ensures the final SG configuration meets the desired benchmarks for each PI. If the design fails to meet these criteria, targeted adjustments are made within the Design Space, ensuring optimized spatial variations in actuation and compliance fields align with task-specific demands before implementation.

To summarize, Algorithm 1 unifies the hierarchical structure and stepwise approach of the optimization framework, aligning each stage with the design objectives and performance indices outlined in the preceding sections.

\begin{algorithm}[t]
\caption{Optimization Framework for Soft Grippers}
\label{alg:OptimizationFramework}

\textbf{Input:} Initial design parameters, target manipulation tasks \\
\textbf{Output:} Optimized multi-fingered soft gripper configuration

\begin{algorithmic}

    \State \textbf{Task Space: Define High-Level Performance Objectives}
    \begin{itemize}
        \item Establish overall performance goals based on manipulation tasks.
        \item Define core performance indices (PIs) for optimization.
        \item Set evaluation criteria for each PI, tailored to task requirements.
    \end{itemize}

    \State \textbf{Motion Space: Translate Objectives into Motion Primitives}
    \begin{itemize}
        \item Identify required motion primitives (e.g., bending, twisting) essential to PIs.
        \item Map each PI to specific movement behaviors that achieve desired task outcomes.
    \end{itemize}

    \State \textbf{Design Space: Optimization Model and Implementation}
    \begin{itemize}
        \item \textbf{Optimization Model}
            \begin{itemize}
                \item Define \textbf{Variables:} geometric parameters, joint articulation, material properties, actuation setups.
                \item Apply \textbf{Constraints:} mechanical, geometric, actuation, kinematic.
                \item Formulate a multi-objective function incorporating the PIs.
            \end{itemize}
        \item \textbf{Optimization Implementation}
            \begin{itemize}
                \item Perform parametric optimization for refined geometry and material properties.
               \item Use topological optimization for structural efficiency and material distribution.
               \item Apply field optimization to achieve spatially varying compliance and actuation control.

            \end{itemize}
    \end{itemize}

    \State \textbf{Sensitivity Analysis:} Determine critical variables impacting PIs.
    \While{Design objectives unmet}
        \State Refine key variables based on sensitivity findings.
        \State Re-evaluate performance indices after adjustments.
    \EndWhile

    \State \textbf{Final Evaluation:} Validate the optimized design against the established PIs.
    \If{Performance criteria unmet}
        \State Reiterate Design Space adjustments as needed.
    \EndIf

\end{algorithmic}
\end{algorithm}

\section{Discussion and Conclusion}

This work introduces a structured, hierarchical approach to the design optimization of soft grippers, addressing the need for systematically balancing multiple performance indices. By structuring the optimization process into Task, Motion, and Design Spaces, the framework ensures alignment between high-level objectives, movement primitives, and design refinements. This layered approach enables SGs to meet practical manipulation demands while avoiding the limitations of conventional methods that prioritize isolated mechanical characteristics over comprehensive performance.

Unlike existing studies that focus on optimizing individual design objectives, this framework adopts a holistic, multi-objective optimization model. By defining and refining dexterity, adaptability, stability, and reachability indices, the framework systematically explores trade-offs, ensuring robust SG designs suited for complex manipulation tasks. This methodology advances soft robotics by promoting balanced performance across all indices, equipping SGs with the capabilities required for adaptive and reliable interactions.

A key consideration in this framework is the choice of multi-objective optimization over weighted single-objective methods. Since performance indices are independent yet competing, predefined weight assignments introduce subjective bias, limiting flexibility in exploring optimal solutions. By leveraging a Pareto-optimal approach, this framework achieves a balanced trade-off across adaptability, dexterity, stability, and reachability, enhancing SG versatility across diverse applications.

Future work will focus on quantitatively defining each performance index, transitioning from conceptual formulations to standardized evaluation metrics for systematic benchmarking \cite{wang2024performance}. Establishing precise metrics will enable more reliable performance comparisons across SG designs. Additionally, refining the Motion Space to establish stronger correlations between indices and movement primitives could further enhance design integration. 
Empirical validation remains a priority, along with further exploration of sensitivity analysis to identify the most influential design parameters. This refinement will enhance optimization efficiency by focusing computational efforts on the most impactful variables, ensuring robust SG performance in real-world applications. 
Finally, defining performance benchmarks and assessment indicators will support broader adoption, facilitating deployment in fields such as healthcare, logistics, and human-machine interaction.

\bibliographystyle{IEEEtran}


\end{document}